\documentclass[10pt,journal,compsoc]{IEEEtran}

\usepackage{epsfig}
\usepackage{graphicx}
\usepackage{amsmath}
\usepackage{amssymb}
\usepackage{subfigure}
\usepackage{multirow}
\usepackage{ragged2e}
 \usepackage{color}
 \usepackage{xcolor}
\usepackage{subfigure}

\usepackage{soul}

\ifCLASSOPTIONcompsoc
  \usepackage[nocompress]{cite}
\else
  \usepackage{cite}
\fi

\ifCLASSINFOpdf
  \else
  \fi

\begin{document}

\title{Reconstruct and Represent Video Contents for Captioning via Reinforcement Learning}
\author{Wei Zhang\qquad Bairui Wang\qquad Lin Ma\qquad  Wei Liu
\IEEEcompsocitemizethanks{\IEEEcompsocthanksitem 
W. Zhang and B. Wang are with the School of Control Science and Engineering, Shandong University, China.  L. Ma, and W. Liu are with Tencent AI Lab, China. (Corresponding author: Lin Ma)\protect\\
E-mail: davidzhang@sdu.edu.cn; \{bairuiwong,forest.linma\}@gmail.com; wl2223@columbia.edu.} }

\IEEEtitleabstractindextext{%
\begin{abstract}
\justifying
In this paper, the problem of describing visual contents of a video sequence with natural language is addressed. Unlike previous video captioning work mainly exploiting the cues of video contents to make a language description, we propose a reconstruction network (RecNet) in a novel encoder-decoder-reconstructor architecture, which leverages both forward (video to sentence) and backward (sentence to video) flows for video captioning. Specifically, the encoder-decoder component makes use of the forward flow to produce a sentence description based on the encoded video semantic features. Two types of reconstructors are subsequently proposed to employ the backward flow and reproduce the video features from local and global perspectives, respectively, capitalizing on the hidden state sequence generated by the decoder. Moreover, in order to make a comprehensive reconstruction of the video features, we propose to fuse the two types of reconstructors together. The generation loss yielded by the encoder-decoder component and the reconstruction loss introduced by the reconstructor are jointly cast into training the proposed RecNet in an end-to-end fashion. Furthermore, the RecNet is fine-tuned by CIDEr optimization via reinforcement learning, which significantly boosts the captioning performance. Experimental results on benchmark datasets demonstrate that the proposed reconstructor can boost  the performance of video captioning consistently.  
\end{abstract}

}

\maketitle

\IEEEdisplaynontitleabstractindextext

\IEEEpeerreviewmaketitle

\section{Introduction}
\label{sec:intro}

\IEEEPARstart{D}{escribing} visual contents with natural language automatically has received increasing attention in both the computer vision and natural language processing communities. It can be applied in various practical applications, such as image and video retrieval \cite{ma2015multimodal,wang2016learning,7535082,song2017quantization,wang2017survey}, answering questions from images~\cite{ma2016learning},
and assisting people who suffer from vision disorders~\cite{Voykinska2016How}.

Previous work predominantly focused on describing still images with natural language \cite{karpathy2014deep,Vinyals_2015_CVPR,vinyals2017show,ren2017deep,jiang2018learning,chen2016sca}. Recently, researchers have striven to generate sentences to describe video contents \cite{xu2015jointly,zhou2018endtoend,li2018jointly,donahue2015long,DBLP:conf/iccv/VenugopalanRDMD15,venugopalan2015translating,pan2016video}. Compared to image captioning~\cite{jiang2018recurrent,chen2018regularizing}, describing videos is more challenging as the amount of information (\textit{e.g.}, objects, scenes, actions, \textit{etc}.) contained in videos is much more sophisticated than that in still images. More importantly, the temporal dynamics within video sequences need to be adequately captured for captioning, besides the spatial content modeling.

\begin{figure}[h]
  \centering
  \includegraphics[width=1\linewidth]{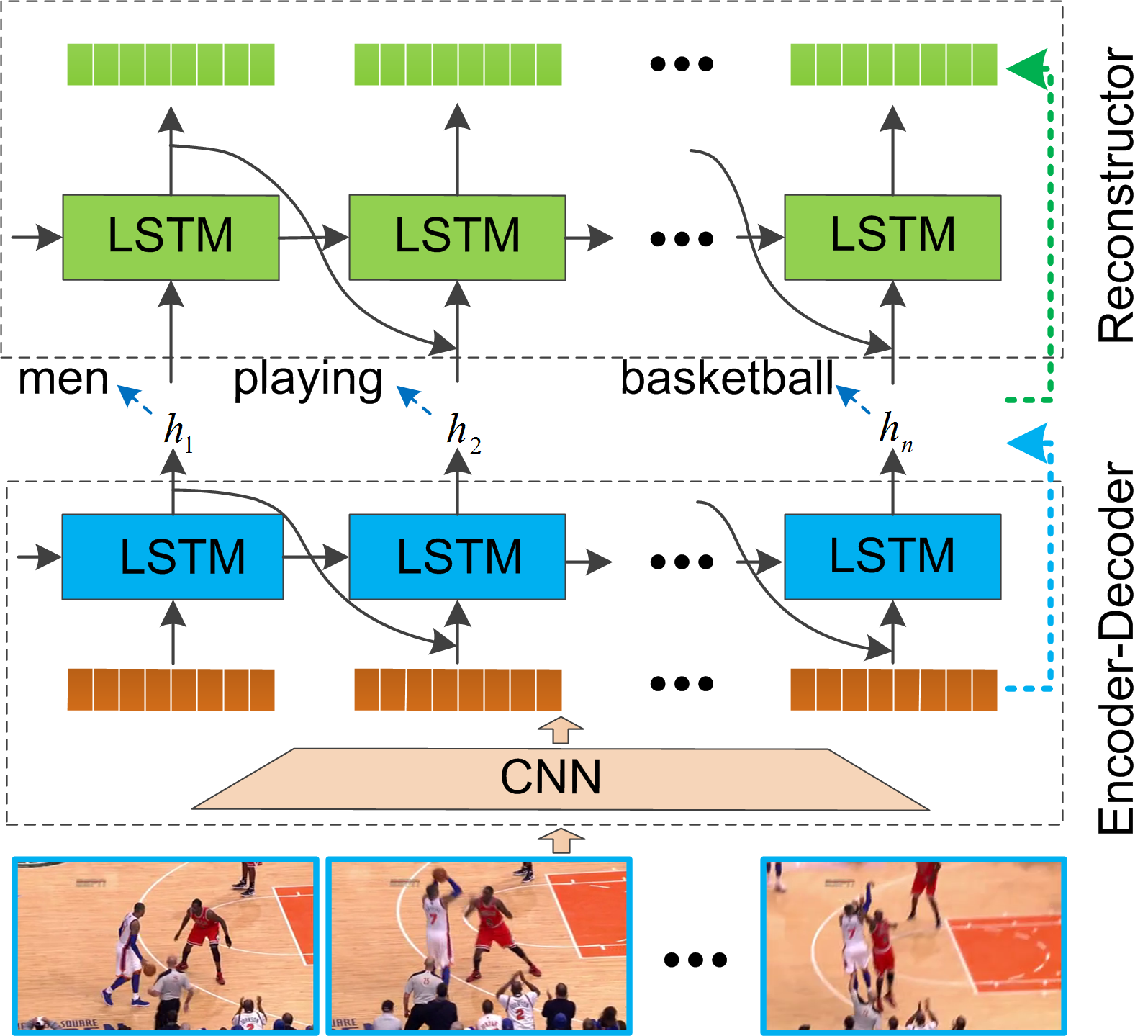}
  \caption{The proposed RecNet with an encoder-decoder-reconstructor architecture. The encoder-decoder relies on the forward flow from video to caption (blue dotted arrow), in which the decoder generates a caption with the frame features yielded by the encoder. The reconstructor, exploiting the backward flow from caption to video (green dotted arrow), takes the hidden state sequence of the decoder as input and reproduces the visual features of the input video.}
  \label{fig1}
\end{figure}

Recently, an encoder-decoder architecture has been widely adopted for video captioning \cite{donahue2015long,ramanishka2016multimodal,jin2016describing,DBLP:conf/iccv/YaoTCBPLC15,gao2017video,song2017hierarchical,pan2016jointly,pan2016video,liu2017video}, as shown in Fig.~\ref{fig1}. However, the encoder-decoder architecture only relies on the forward flow (video to sentence), but does not consider the information from sentence to video, named as backward flow. Usually the encoder is a convolutional neural network (CNN) capturing the image structure to yield its semantic representation. For a given video sequence, the yielded semantic representations by a CNN are further fused together to exploit the video temporal dynamics and generate the video representation. The decoder is usually a long short-term memory (LSTM)~\cite{hochreiter1997long} or a gated recurrent unit (GRU)~\cite{cho2014properties}, which is popular in processing sequential data \cite{zhang2017learning}.
LSTM and GRU generate the sentence fragments one by one, and ensemble them to form one sentence. The semantic information from target sentences to source videos is never included.
Actually, the backward flow can be yielded by a dual learning mechanism that has been introduced into neural machine translation (NMT)~\cite{tu2017neural,he2016dual} and image segmentation~\cite{luo2017deep}. This mechanism reconstructs a source from a target when the target is achieved, and demonstrates that the backward flow from target to source improves the performance.

To well exploit the backward flow, we refer to the idea of dual learning and propose an encoder-decoder-reconstructor architecture shown in Fig.~\ref{fig1}, denoted as RecNet, to address video captioning. Specifically, the encoder fuses the video frame features together to exploit the video temporal dynamics and generate the video representation, based on which the decoder generates the corresponding sentences. The reconstructor, realized by one LSTM, leverages the backward flow (sentence to video). That is, the reconstructor reproduces the video information based on the hidden state sequence generated by the decoder. {Such an encoder-decoder-reconstructor can be viewed as a dual learning architecture, where video captioning is the primal task and reconstruction behaves as its dual task. In the dual task, a reconstruction loss measuring the difference between the reconstructed and original visual features is additionally used to train the primal task and optimize the parameters of the encoder and decoder. With such a reconstructor, the decoder is encouraged to embed more information from the input video sequence. Therefore, the relationship between the video sequence and caption can be further enhanced. And the decoder can generate more semantically correlated sentences to describe the visual contents of the video sequences, yielding significant performance improvements.} As such, the proposed reconstructor, which further helps finetune the parameters of the encoder and decoder, is expected to bridge the semantic gap between the video and sentence.

Moreover, the reconstructor can help mitigate the discrepancy, also referred to as the exposure bias~\cite{ranzato2015sequence}, between the training and inference processes, which widely exists in RNNs for the captioning task. The proposed reconstructor can help regularize the transition dynamics of RNNs, as the dual information captured by the RecNet provides a complementary view to the encoder-decoder architecture. As such, the reconstructor can help alleviate the exposure bias between training and inference and mitigate the discrepancy, as will be demonstrated in Sec.~\ref{sec:curve_and_exposure}. 

Besides, we intend to directly train the captioning models guided by evaluation metrics, such as BLEU and CIDEr, instead of the conventionally used cross-entropy loss. However, these evaluation metrics are discrete and non-differentiable, which makes it difficult to optimize using traditional methods.
The self-critical sequence  training~\cite{rennie2016self} is an excellent REINFORCE-based algorithm, 
specializing in processing the discrete and non-differentiable variables. In this paper, we resort to the self-critical sequence training algorithm to further boost the performance of the RecNet on the video captioning task.

{
To summarize, our main contributions of this work are as follows. We propose a novel reconstruction network and build an end-to-end encoder-decoder-reconstructor architecture to exploit both the forward (video to sentence) and backward (sentence to video) flows for video captioning. Two types of reconstructors are customized to recover the global and local structures of the video, respectively. Moreover, a joint model is presented to reconstruct both the global and local structures simultaneously for further improving the reconstruction of the video representation. Extensive results obtained by cross-entropy training and self-critical sequence training~\cite{rennie2016self} on benchmark datasets indicate that the backward flow is well exploited by the proposed reconstructors, and considerable gains on video captioning are achieved. Besides, ablation studies show that the proposed reconstructor can help regularize the transition dynamics of RNNs, thereby mitigating the discrepancy between training and inference processes.}

\section{Related Work}
In this section, we first introduce two types of video captioning: template-based approaches \cite{kojima2002natural,guadarrama2013youtube2text,rohrbach2013translating,rohrbach2014coherent,xu2015jointly} and sequence learning approaches \cite{DBLP:conf/iccv/YaoTCBPLC15,DBLP:conf/iccv/VenugopalanRDMD15,venugopalan2015translating,donahue2015long,ramanishka2016multimodal,jin2016describing,zhang2016automatic,pan2016jointly,pan2016video,shen2017weakly,liu2017video,wang2018bidirectional}, and then introduce the application of dual learning.

\subsection{Template-based Approaches}
Template-based methods first define some specific rules for language grammar, and then parse the sentence into several components such as subject, verb, and object. The obtained sentence fragments are associated with words detected from the visual content to produce the final description about an input video with predefined templates. For example, a concept hierarchy of actions was introduced to describe human activities in \cite{kojima2002natural}, while a semantic hierarchy was defined in \cite{guadarrama2013youtube2text} to learn the semantic relationship between different sentence fragments. In \cite{rohrbach2013translating}, the conditional random field (CRF) was adopted to model the connections between objects and activities of the visual input and generate the semantic features for description. Besides, Xu \textit{et al.} proposed a unified framework consisting of a semantic language model, a deep video model, and a joint embedding model to learn the association between videos and natural sentences \cite{xu2015jointly}. However, as stated in \cite{pan2016video}, the aforementioned approaches highly depend on  predefined templates and are thus limited by the fixed syntactical structure, which is inflexible for sentence generation.

\begin{figure*}
  \centering
  \includegraphics[width=1\linewidth]{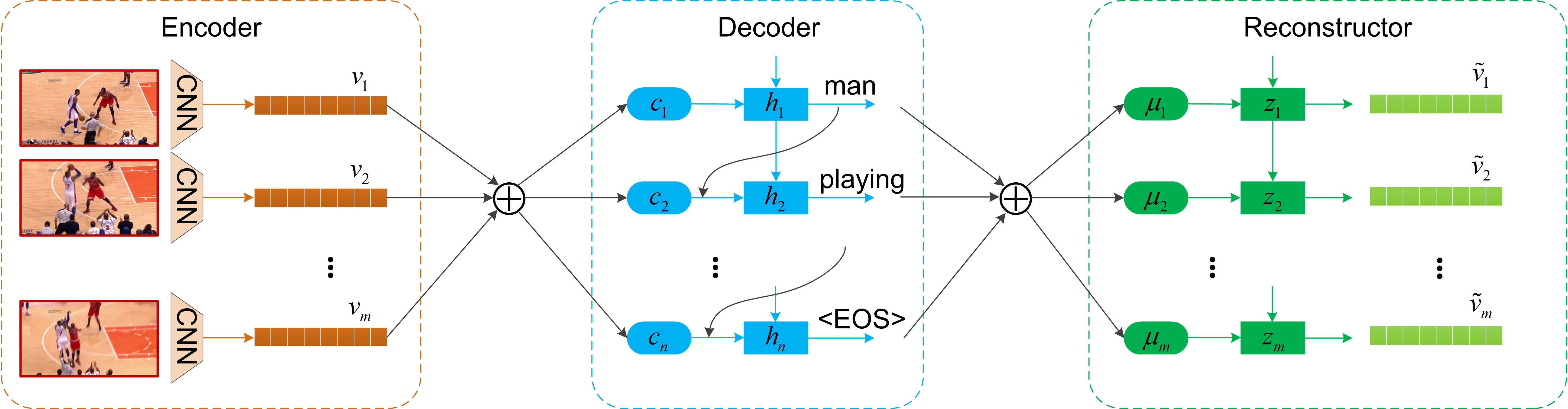}
  \caption{The proposed RecNet consists of three parts: the CNN-based encoder which extracts the semantic representations of the video frames, the LSTM-based decoder which generates natural language for visual content description, and the reconstructor which exploits the backward flow from caption to visual contents to reproduce the frame representations.} 
  \label{fig2}
\end{figure*}

\subsection{Sequence Learning Approaches}
Compared with the template-based methods, the sequence learning approaches aim to directly produce a sentence description about the visual input with more flexible syntactical structures. For example, in \cite{venugopalan2015translating}, a video representation was obtained by averaging each frame feature extracted by a CNN, and then fed to LSTMs for sentence generation. In \cite{pan2016jointly}, the relevance between video context and sentence semantics was considered as a regularizer in the LSTM. However, since simple mean pooling is used, the temporal dynamics of the video sequence are not adequately  addressed. Yao \textit{et al.} introduced an attention mechanism to assign weights to the features of each frame and then fused them based on the attentive weights \cite{DBLP:conf/iccv/YaoTCBPLC15}. Venugopalan \textit{et al.} proposed S2VT \cite{DBLP:conf/iccv/VenugopalanRDMD15}, which includes the temporal information with optical flow and employs LSTMs in both the encoder and decoder. To exploit both temporal and spatial information, Zhang and Tian proposed a two-stream encoder comprised of two 3D CNNs \cite{DBLP:journals/corr/TranBFTP14,DBLP:conf/cvpr/KarpathyTSLSF14} and one parallel fully connected layer to learn the features from the frames \cite{zhang2016automatic}. Besides, Pan \textit{et al.} proposed a transfer unit to model the high-level semantic attributes from both images and videos, which are rendered as the complementary knowledge to video representations for boosting sentence generation \cite{pan2016video}. 

More recently, reinforcement learning has shown benefits on video captioning tasks. Pasunuru and Bansal employed reinforcement learning to directly optimize the CIDEt scores (a variant metric of CIDEr) and achieved state-of-the-art results on the MSR-VTT dataset~\cite{pasunuru2017reinforced}. Wang \textit{et al.} proposed a hierarchical reinforcement learning framework, where a manager guides a worker to generate semantic segments about activities to produce more detailed descriptions. 

\subsection{Dual Learning Approaches}
As far as we know, dual learning mechanism has not been employed in video captioning but widely used in NMT~\cite{tu2017neural,he2016dual,xia2017dual}. In~\cite{tu2017neural}, the source sentences are reproduced from the target side hidden states, and the accuracy of reconstructed source provides a constraint for the decoder to embed more information of source language into target language. In~\cite{he2016dual}, the dual learning is employed to train model of inter-translation of English-French, and obtains significant improvement on tasks of English to French and French to English.

In this paper, our proposed RecNet can be regarded as a sequence learning method. However, unlike the above conventional encoder-decoder models which only depend on the forward flow from video to sentence, RecNet can also benefit the backward flow from sentence to video. By fully considering the bidirectional flows between video and sentence, RecNet is capable of further boosting the video captioning.
Besides, it is worth noting that this work is an extended version of~\cite{wang2018reconstruction}. The main improvements of this version are elaborated as follows:
First, this work takes one step forward and presents a new reconstruction model, named as RecNet$_{g+l}$, which considers both global and local structures to further improve the reconstruction of the video representation. Second, the exposure bias between training and inference processes is studied in this work. We demonstrate that the proposed reconstructor can help regularize the transition dynamics of the decoder, and mitigate the discrepancy between training and inference processes. Besides, more ablation studies on reconstructors are conducted, including training with the self-critical algorithm, visualization of the hidden states of the decoder, and curves of the training losses and metrics used to examine how well the reconstrutor works. Also, an additional dataset ActivityNet~\cite{caba2015activitynet} is included to further verify the effectiveness of the proposed reconstructor.

\section{Architecture}
We propose a novel RecNet with an encoder-decoder-reconstructor architecture for video captioning, which works in an end-to-end manner. The reconstructor imposes one constraint that the semantic information of one source video can be reconstructed from the hidden state sequence of the decoder. The encoder and decoder are thus encouraged to embed more semantic information about the source video. As illustrated in Fig.~\ref{fig2}, the proposed RecNet consists of three components, specifically the encoder, the decoder, and the reconstructor:
 \begin{itemize}
   \item \textbf{Encoder.} Given one video sequence, the encoder yields the semantic representation for each video frame.
   \item \textbf{Decoder.} The decoder decodes the corresponding representations generated by the encoder into one caption describing the video content.
   \item \textbf{Reconstructor.} The reconstructor takes the intermediate hidden state sequence of the decoder as input, and reconstructs the video global or local structure. 
 \end{itemize}

Moreover, our designed reconstructor can collaborate with different classical encoder-decoder architectures for video captioning. Our proposed reconstructor can be built on top of any classical encoder-decoder models for video captioning.
In this paper, we employ the attention-based video captioning \cite{DBLP:conf/iccv/YaoTCBPLC15} and S2VT \cite{DBLP:conf/iccv/VenugopalanRDMD15} as the classical encoder-decoder models. We first briefly introduce the encoder-decoder model for video captioning. Afterward, the proposed reconstructors with different architectures are described.

\subsection{Encoder-Decoder}
\label{ed}
The aim of video captioning is to generate one sentence $\mathbf{S}=\{\mathbf{s}_1,\mathbf{s}_2, \dots ,\mathbf{s}_n\}$ to describe the content of one given video $\mathbf{V}$. Classical encoder-decoder architectures directly model the captioning generation probability word by word:
\begin{equation}
\label{eq:XE_encoder-decoder}
P( \mathbf{S}|\mathbf{V} )=\prod_{i=1}^{n}  P\left ( \mathbf{s}_{i} | \mathbf{s}_{< i},\mathbf{V}; \theta \right ),
\end{equation}
where $\theta$ keeps the parameters of the encoder-decoder model. $n$ denotes the length of the sentence, and $\mathbf{s}_{< i}$ (\textit{i.e.}, $\{ \mathbf{s}_1, \mathbf{s}_2, \dots, \mathbf{s}_{i-1} \}$) denotes the generated partial caption.

\vspace{5pt}
\noindent\textbf{Encoder.} {To generate reliable captions, visual features need to be extracted to capture the high-level semantic information about the video.
Previous methods usually rely on CNNs, such as AlexNet \cite{venugopalan2015translating}, GoogleNet \cite{DBLP:conf/iccv/YaoTCBPLC15}, and VGG19 \cite{xu2016msr} to
encode each video frame into a fixed-length representation with the high-level semantic information. By contrast, in this work, considering a deeper network is more plausible for feature extraction, we advocate using Inception-V4 \cite{DBLP:conf/aaai/SzegedyIVA17} as the encoder. In this way, the given video sequence is encoded as a sequential representation $\mathbf{V}=\{\mathbf{v}_1, \mathbf{v}_2, \dots, \mathbf{v}_m\}$, where $m$ denotes the total number of the video frames.}

\vspace{5pt}
\noindent\textbf{Decoder.} Decoder aims to generate the caption word by word based on the video representation.
LSTMs with the capabilities of modeling long-term temporal dependencies are used to decode video representation to video captions word by word. To further exploit the global temporal information of videos, a temporal attention mechanism \cite{DBLP:conf/iccv/YaoTCBPLC15} is employed to encourage the decoder to select the key frames/elements for captioning.

During the captioning process, the $i_{th}$ word prediction is generally made by the LSTM:
\begin{gather}
\label{eq:e_d_lstm}
\begin{split}
  x_i = \textrm{Linear}\big(f(\mathbf{s}_{i-1},h_i,c_i;\theta)\big),\\
  P\left ( \mathbf{s}_{i} | \mathbf{s}_{< i},\mathbf{V}, \theta \right ) \sim \textrm{Softmax}(x_i),
\end{split}
\end{gather} 
where $f$ represents the LSTM activation function, $h_i$ is the $i_{th}$ hidden state computed in the LSTM, and $c_i$ denotes the $i_{th}$ context vector computed with the temporal attention mechanism, which is used to decode the $i_{th}$ word. The temporal attention mechanism is used to assign weight $\alpha_{j}^{i}$ to the representation of each frame $\left \{ \mathbf{v}_{1},\mathbf{v}_{2}, \dots,\mathbf{v}_{m} \right \}$ at the time step $i$ as follows:
\begin{equation}
\label{eq:3}
c_i=\sum_{j=1}^{m}\alpha_{j}^{i}\mathbf{v}_{j},
\end{equation}
where $m$ denotes the number of the video frames and $\sum_{j=1}^{m}{\alpha_j^i}=1$. 
In order to obtain the attentive weight $\alpha_j^i$ at the $i_{th}$ time step for the $j_{th}$ video frame representation, we follow the traditional way in~\cite{DBLP:conf/iccv/YaoTCBPLC15} to calculate the relevance score $e^i_j$ for the $j_{th}$ frame representation with respect to the hidden state $h_{i-1}$:
\begin{equation}
\label{eq:3_1}
e^i_j = \text{w}_{\alpha}^{\top}\tanh(\text{w}_{vd}v_j + \text{w}_{hd}h_{i-1} + b_d),
\end{equation}
where $\text{w}_{\alpha}^{\top}$, $\text{w}_{vd}$, $\text{w}_{hd}$, and $b_d$ are the learnable parameters. 
The attentive weight $\alpha_j^i$ is thereby obtained by:
\begin{equation}
\label{eq:3_2}
\alpha_j^i = \frac{\exp(e_j^i)}{\sum_{k=1}^m \exp(e_k^i)}.
\end{equation}
The attentive weight $\alpha_j^i$ reflects the relevance between the $j_{th}$ frame representation and the hidden state $h_{i-1}$ at the $(i-1)_{th}$ time step. The larger $\alpha_j^i$, the more relevant the $j_{th}$ video frame representation $v_j$ is to $h_{i-1}$, which allows the decoder to focus more on the corresponding video frames to generate the word at the next time step.

\subsection{Reconstructor}
As shown in Fig.~\ref{fig2}, the proposed reconstructor is built on  top of the encoder-decoder, which is expected to reproduce the video from the hidden state sequence of the decoder. However, due to the diversity and high dimension of the video frames, directly reconstructing the video frames seems to be intractable. Therefore, in this paper, the reconstructor aims at reproducing the sequential video frame representations generated by the encoder, with the hidden states $\mathbf{H}=\left \{ h_{1},h_{2},...,h_{n} \right \}$ of the decoder as input. The benefits of such a structure are two-fold. First, the proposed encoder-decoder-reconstructor architecture can be trained in an end-to-end fashion. Second, with such a reconstruction process, the decoder is encouraged to embed more information from the input video sequence. Therefore, the relationship between the video sequence and caption can be further enhanced, which is expected to improve the video captioning performance.

In practice, the reconstructor is realized by LSTMs. Two different architectures are customized to summarize the hidden states of the decoder for video feature reproduction. More specifically, one focuses on reproducing the global structure of the provided video, while the other pays more attentions to the local structure by selectively attending to the hidden state sequence.
Moreover, to comprehensively reconstruct the video sequence, we propose a new architecture to reconstruct both the global and local structures of the video features.

\subsubsection{Reconstructing Global Structure}
\begin{figure}
  \centering
  \includegraphics[width=1\linewidth]{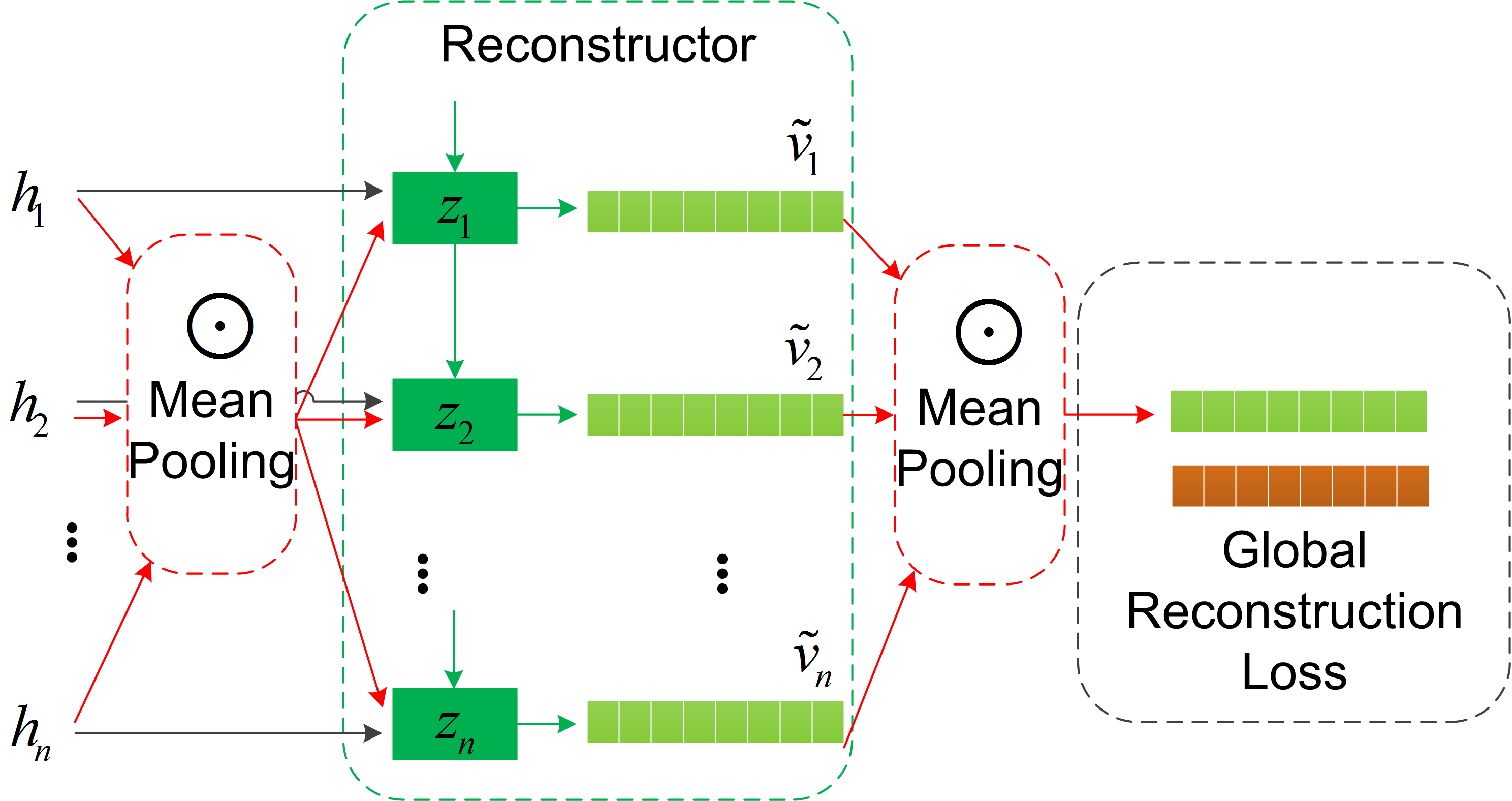} 
  \caption{An illustration of the proposed reconstructor that reproduces the global structure of the video sequence. {The left mean pooling is employed to summarize the hidden states of the decoder for the global representation of the caption. The reconstructor aims to reproduce the feature representation of the whole video by mean pooling (the right one) using the global representation of the caption as well as the hidden state sequence of the decoder.}}
  \label{fig3}
\end{figure}

The architecture for reconstructing the global structure of the video sequence is illustrated in Fig.~\ref{fig3}.
The whole sentence is fully considered to reconstruct the video global structure. Therefore, besides the hidden state $h_t$ at each time step, the global representation characterizing the semantics of the whole sentence is also taken as the input at each step.
Several methods like LSTM and multiple-layer perception can be employed to fuse the hidden sequential states of the decoder to generate the global representation. Inspired by \cite{DBLP:conf/iccv/VenugopalanRDMD15}, the mean pooling strategy is performed on the hidden states of the decoder to yield the global representation of the caption:
\begin{equation}
\phi \left ( \mathbf{H} \right )=\frac{1}{n}\sum_{i=1}^{n}h_{i} ,
\end{equation}
where $\phi \left ( \cdot  \right )$ denotes the mean pooling process, which yields a vector representation $\phi \left ( \mathbf{H} \right )$ with the same size as $h_{i}$. Thus, the LSTM unit of the reconstructor is further modified as:
\begin{align}
\label{eq_lstm_mp}
\begin{split}
  \begin{pmatrix}i_t \\ f_t \\ o_t \\ g_t \end{pmatrix} &=
  \begin{pmatrix} \sigma \\ \sigma \\ \sigma \\ \tanh \end{pmatrix}
  \mathbf{T}
  \begin{pmatrix} h_t \\ z_{t-1}\\ \phi \left ( \mathbf{H} \right )\end{pmatrix}, \\
  m_t &= f_t \odot m_{t-1} + i_t \odot g_t, \\
  z_t &= o_t \odot \tanh(m_t),
\end{split}
\end{align}
where $i_t$, $f_t$, $m_t$, $o_t$, and $z_t$ denote the input, forget, memory, output, and hidden states of each LSTM unit, respectively. $\sigma$ and $\odot$ denote the logistic sigmoid activation and the element-wise multiplication, respectively.

To reconstruct the video global structure from the hidden state sequence produced by the encoder-decoder, the global reconstruction loss is defined as:
\begin{equation}
\label{eq_global_loss}
\mathcal{L}^g_{rec} = \psi\big(\phi(\mathbf{V}),\phi(\mathbf{Z})\big),
\end{equation}
where $\phi(\mathbf{V})$ denotes the mean pooling process on the video frame features, yielding the ground-truth global structure of the input video sequence. $\phi(\mathbf{Z})$ works on the hidden states of the reconstructor, indicating the global structure recovered from the captions. The reconstruction loss is measured by $\psi(\cdot)$, which is simply chosen as the Euclidean distance.

\begin{figure}
  \centering
  \includegraphics[width=1\hsize]{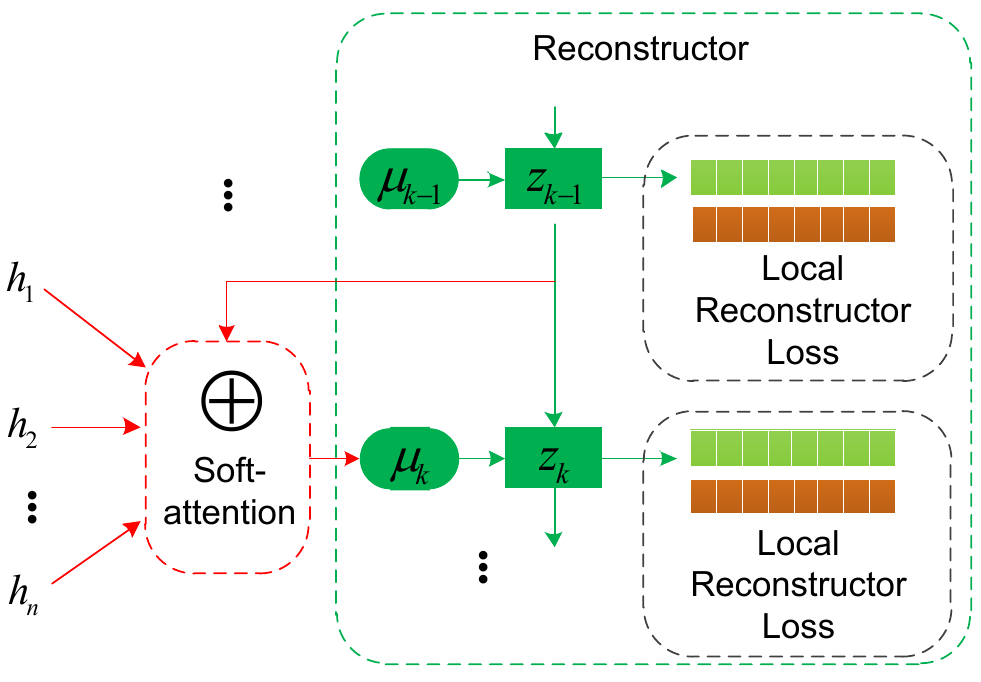} 
  \caption{An illustration of the proposed reconstructor that reproduces the local structure of the video sequence. The reconstructor works on the hidden states of the decoder by selectively adjusting the attentive weights, and reproduces the feature representation frame by frame.}
  \label{fig4}
\end{figure}

\subsubsection{Reconstructing Local Structure}

\begin{figure*}
  \centering
  \includegraphics[width=1\hsize]{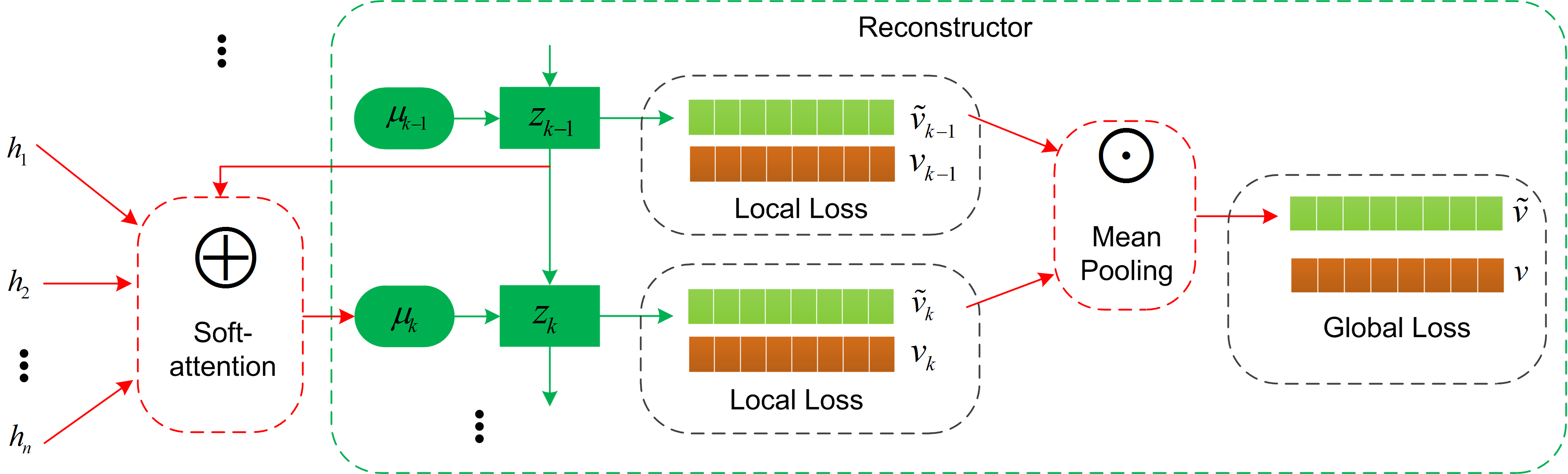} 
  \caption{An illustration of the proposed reconstructor that reproduces both the global structure and local structure of the video sequence. The reconstructor works on the hidden states of the decoder by selectively adjusting the attentive weights, and reproduces the feature representation frame by frame. Mean pooling is conducted to summarize the reproduced feature representation sequence to yield the representation of the whole video.} 
  \label{fig:global_local}
\end{figure*}

\label{sec:local_struct}
The aforementioned reconstructor aims to reproduce the global representation for the whole video sequence, while neglects the local structures in each frame. In this subsection, we propose to learn and preserve the temporal dynamics by reconstructing each video frame as shown in Fig.~\ref{fig4}. Differing from the global structure estimation, we intend to reproduce the feature representation of each frame from the key hidden states of the decoder selected by the attention strategy \cite{DBLP:journals/corr/BahdanauCB14,DBLP:conf/iccv/YaoTCBPLC15}:

\begin{equation}
\label{eq:context}
\mu_t=\sum_{i=1}^{n}\beta_{i}^{t}h_{i},
\end{equation}
where $\sum_{i=1}^{n}\beta_i^t=1$ and $\beta_i^t$ denotes the weight computed for the $i_{th}$ hidden state from the decoder at time step $t$ by the attention mechanism. Here $\beta_i^t$ measures the relevance of the $i_{th}$ hidden state $h_i$ of the decoder given the previously reconstructed frame representations $z_{t-1}$. 
Similar to Eqs.~(\ref{eq:3_1}) and (\ref{eq:3_2}), to calculate the attentive weight $\beta_i^t$, the relevance score for the $i_{th}$ hidden state from the decoder with respect to the previous hidden state $z_{t-1}$ in the reconstructor is first computed by:
\begin{equation}
\label{eq:context_1}
\begin{split}
    e^t_i &= \text{w}_{\beta}^{\top}\tanh(\text{w}_{hr}h_i + \text{w}_{zr}z_{t-1} + b_r), \\
    \beta_i^t &= \frac{\exp(e_i^t)}{\sum_{k=1}^n \exp(e_k^t)},
\end{split}
\end{equation}
where $\text{w}_{\beta}^{\top}$, $\text{w}_{hr}$, $ \text{w}_{zr}$, and $b_r$ are the learnable parameters, and $n$ denotes the total number of hidden states from the decoder.

Such a strategy encourages the reconstructor to work on the hidden states selectively by adjusting the attentive weight $\beta_i^t$ and yield the context information $\mu_t$ at each time step as in Eq.~(\ref{eq:context}). As such, the proposed reconstructor can further exploit the temporal dynamics and the word compositions across the whole caption. The LSTM unit is thereby reformulated as:

\begin{align}
\label{eq_lstm_sa}
\begin{split}
  \begin{pmatrix}i_t \\ f_t \\ o_t \\ g_t \end{pmatrix} &=
  \begin{pmatrix} \sigma \\ \sigma \\ \sigma \\ \tanh \end{pmatrix}
  \mathbf{T}
  \begin{pmatrix} \mu_t \\ z_{t-1} \end{pmatrix}.
\end{split}
\end{align}

Differing from the global structure recovery step in Eq.~(\ref{eq_lstm_mp}), the dynamically generated context $\mu_t$ is taken as the input other than the hidden state $h_t$ and its mean pooling representation $\phi \left ( \mathbf{H} \right )$. Moreover, instead of directly generating the global mean representation of the whole video sequence, we propose to produce the feature representation frame by frame. 
The reconstruction loss is thereby defined as:
\begin{equation}
\label{eq_local_loss}
\mathcal{L}^l_{rec} = \frac{1}{m}\sum_{j=1}^{m}\psi(z_{j},\mathbf{v}_{j}).
\end{equation}

\subsubsection{Reconstructing both Global and Local Structures}
\label{combine_global_local}
In this subsection, we step further and intend to reconstruct both global and local structures of the video sequence. The architecture is illustrated in Fig.~\ref{fig:global_local}. 

Different from the aforementioned two methods, we first reconstruct the feature representation $\{z_1,z_2,\dots,z_{m}\}$ of each frame with the local information of the input video.
After that, mean pooling is conducted on the reproduced frame representation and the global representation of the video is yielded. The global reconstruction loss and the local reconstruction loss are jointly considered as follows:
\begin{equation}
\label{eq:global+local}
\mathcal{L}^{g+l}_{rec} = \psi\big(\phi(\mathbf{Z}), \phi(\mathbf{V}) \big) + \frac{1}{m}\sum_{j=1}^{m}\psi(z_{j},\mathbf{v}_{j}),
\end{equation}
where the first part denotes the global reconstruction loss computed by Eq.~(\ref{eq_global_loss}), while the second part is the local reconstruction loss computed by Eq.~(\ref{eq_local_loss}). 

Such an architecture is designed to reproduce both the global and local information. It is expected to comprehensively exploit the backward information flow and further boost the performance of video captioning.

\subsection{Training}

\subsubsection{Training Encoder-Decoder}

Traditionally, the encoder-decoder model can be jointly trained by minimizing the {negative log likelihood} to produce the correct description sentence given the video as follows:
\begin{equation}
\label{eq_encoder_decoder_obj}
\min_\theta\sum_{i=1}^N\left\{-\log   P\left ( \mathbf{S}^{i} | \mathbf{V}^i; \theta \right ) \right\}.
\end{equation}

Moreover, in order to directly optimize the metrics instead of the cross entropy loss, we consider the video contents and words as "environment", and the encoder-decoder model as "agent" which interacts with the environment. At each step $t$ for agent LSTMs, the policy $\pi_{\theta}$ takes "action" to predict a word followed by the updating of "state", which denotes the hidden states and cell states of LSTMs. When a sentence is generated, the metric score is computed and regarded as the "reward" (here the CIDEr score is taken as the reward), which we intend to optimize by minimizing the negative expected reward:
\begin{equation}
\label{eq:RL_loss_origin}
\mathcal{L}^{RL}_{ED}(\theta) = -\mathbb{E}_{\mathbf{S}^{i}\sim\pi_{\theta}}[ r(\mathbf{S}^{i})],
\end{equation}
where $r(\mathbf{S}^{i})$ denotes the reward of the word sequence sampled from the model. 
The gradient of the non-differentiable loss $\mathcal{L}^{RL}_{ED}$ can be obtained by Eq.~(\ref{eq:RL_loss_origin}) using the REINFORCE algorithm. 
\begin{equation}
\label{eq:gradient_RL_loss_origin}
\triangledown_{\theta}\mathcal{L}^{RL}_{ED} = -\mathbb{E}_{\mathbf{S}^{i}\sim\pi_{\theta}}[r(\mathbf{S}^{i})\cdot \triangledown_{\theta}\log\pi_{\theta}(\mathbf{S}^{i}) ].
\end{equation}  

In fact $\mathcal{L}^{RL}_{ED}$ is typically estimated with a single sample from $\pi_{\theta}$. Therefore, for each sample, Eq.~(\ref{eq:gradient_RL_loss_origin}) can be represented as:
\begin{equation}
\label{eq:gradient_RL_loss}
\triangledown_{\theta}\mathcal{L}^{RL}_{ED} \approx  - r(\mathbf{S}^{i})\cdot \triangledown_{\theta}\log\pi_{\theta}(\mathbf{S}^{i}).
\end{equation}  

To deal with the high variance of the gradient estimate with a single sample, we also add a baseline reward into the reward-based loss to generalize the policy gradient given by REINFORCE, which could reduce the variance without changing the expected gradient:
\begin{equation}
\label{eq:gradient_RL_loss_baseline}
\triangledown_{\theta}\mathcal{L}^{RL}_{ED} \approx  - (r(\mathbf{S}^{i}) - \mathit{b})\cdot \triangledown_{\theta}\log\pi_{\theta}(\mathbf{S}^{i}),
\end{equation}
where $\mathit{b}$ denotes the baseline reward. 

According to the chain rule, the gradient of the loss function can be calculated as:
\begin{equation}
\label{eq:chain_relu}
\triangledown_{\theta}\mathcal{L}^{RL}_{ED} = \sum_{i=1}^{n}\frac{\partial \mathcal{L}^{RL}_{ED}}{\partial x_i} \frac{\partial x_i}{\partial \theta},
\end{equation}
where $x_i$ is the input to the \textit{Softmax} layer in Eq.~(\ref{eq:e_d_lstm}). The estimation of $\frac{\partial \mathcal{L}^{RL}_{ED}}{\partial x_i}$ with the baseline is given by~\cite{zaremba2015reinforcement} as follows:
\begin{equation}
\label{eq:RL_last}
\frac{\partial \mathcal{L}^{RL}_{ED}}{\partial x_i} \approx \big(r(\mathbf{S})-\mathit{b}\big) \big(\pi_{\theta}(s_i | h_i)-1_{s_i} \big).
\end{equation}

Usually the baseline is estimated by a trainable network~\cite{ranzato2015sequence}, which significantly increases the computational complexity. To handle such a drawback, Rennie~\textit{et al.} proposed a self-critical sequence training method~\cite{rennie2016self}, and the baseline is considered as the CIDEr score of the sentence, which is generated by the current encoder-decoder model under the inference mode. It proved to be more effective for training, as such a scheme not only brings a lower gradient variance, but also avoids learning an estimate of expected future rewards as the baseline. Hence, in this paper, we employ the self-critical sequence training method and take the metric score of the sentence $\hat{\mathbf{S}}$ greedily decoded by the current model with the inference algorithm as the baseline, i.e., $\mathit{b} = r(\hat{\mathbf{S}})$. As such, by replacing the baseline $b$ with $r(\hat{\mathbf{S}})$, Eq.~(\ref{eq:RL_last}) can be rewritten as:

\begin{equation}
\label{eq:RL_last_2}
\frac{\partial \mathcal{L}^{RL}_{ED}}{\partial x_i} \approx \big(r(\mathbf{S})-r(\hat{\mathbf{S}})\big) \big(\pi_{\theta}(s_i | h_i)-1_{s_i} \big).
\end{equation}
Hence, if a sample has a reward $r(\mathbf{S})$ higher than baseline $b$, the gradient is negative and such a distribution is encouraged by increasing the probability of the corresponding word. Similarly, the sample distribution that has a low reward is discouraged.

\subsubsection{Training Encoder-Decoder-Reconstructor}

Formally, we train the proposed encoder-decoder-reconstructor architecture by minimizing the whole loss defined in Eq.~(\ref{eq_full_loss}), which involves both the forward (video-to-sentence) likelihood and the backward (sentence-to-video) reconstruction loss: 
\begin{equation}
\label{eq_full_loss}
\begin{split}
\mathcal{L}(\theta,\theta_{rec}) = \sum_{i=1}^N & \Big( \underbrace{ \big[ -\log P\left ( \mathbf{S}^{i} | \mathbf{V}^i; \theta \right ),\ \mathcal{L}^{RL}_{ED} \big]}_{\text{encoder-decoder}}\\
&+ \lambda \underbrace{\mathcal{L}_{rec}(\mathbf{V}^i,\mathbf{Z}^i;\theta_{rec})}_{\text{reconstructor}}   \Big).
\end{split}
\end{equation} 

The loss for the encoder-decoder can be either the traditional cross entropy loss or the reinforcement loss $\mathcal{L}^{RL}_{ED}$. The reconstruction loss $\mathcal{L}_{rec}(\mathbf{V}^i,\mathbf{Z}^i;\theta_{rec})$ can be set as the global loss in Eq.~(\ref{eq_global_loss}), the local loss in Eq.~(\ref{eq_local_loss}), or the joint global and local loss in Eq.~(\ref{eq:global+local}). The hyper-parameter $\lambda$ is introduced to seek a trade-off between the encoder-decoder and the reconstructor.

The training of our proposed RecNet model proceeds in two stages. First, we rely on the forward likelihood 
to train the encoder-decoder component of the RecNet, which is terminated by the early stopping strategy. Afterward, the reconstructor and the backward reconstruction loss $\mathcal{L}_{rec}(\theta_{rec})$ are introduced. We use the whole loss defined in Eq.~(\ref{eq_full_loss}) to jointly train the reconstructor and fine-tune the encoder-decoder. For the reconstructor, the reconstruction loss is calculated using the hidden state sequence generated by the LSTM units in the reconstructor as well as the video frame feature sequence.

\subsection{RecNet vs. Autoencoder}
The whole architecture of the RecNet can be regarded as one autoencoder. Specifically, the encoder-decoder framework acts as the ``encoder'' component in the autoencoder, and its aim is to generate the fluent sentences describing the video contents. The reconstructor performs as the ``decoder'', and its aim is to ensure the semantic correlations between the generated caption and the input video sequence. Also, the reconstruction losses defined in Eqs.~(\ref{eq_global_loss}),~(\ref{eq_local_loss}), and~(\ref{eq:global+local}) act similarly to the reconstruction error defined in the autoencoder, which can further guarantee that the model can learn effective feature representation from the input video sequence \cite{2006Sci...313..504H}. Compared to the encoder-decoder, the proposed reconstructor is able to exploit the backward flow from sentence to video, and acts as one regularizer to make the encoder-decoder component produce captions semantically correlated to input video sequences. Consequently, the video captioning performance can be improved further.

\section{Experimental Results}

In this section, we evaluate the proposed RecNet on the video captioning task over the benchmark datasets such as Microsoft Research video to text (MSR-VTT)~\cite{xu2016msr}, Microsoft Research Video Description Corpus (MSVD)~\cite{chen2011collecting}, and ActivityNet Captions (ActivityNet)~\cite{caba2015activitynet}. We compute the traditional evaluation metrics, namely BLEU-4~\cite{papineni2002bleu}, METEOR~\cite{banerjee2005meteor}, ROUGE-L~\cite{lin2004rouge}, and CIDEr~\cite{vedantam2015cider} with the codes released on the Microsoft COCO evaluation server\footnote{https://github.com/tylin/coco-caption}~\cite{chen2015microsoft} and ActivityNet evaluation server\footnote{https://github.com/ranjaykrishna/densevid\_eval}. In the following, we first introduce the benchmark datasets and the implementation details of the proposed method. Then, the experimental results are provided and discussed.

\subsection{Datasets and Implementation Details}
\label{sec:details}
\subsubsection{Datasets}
\textbf{MSR-VTT.} It is the largest dataset for video captioning so far in terms of the number of video-sentence pairs and the vocabulary size. In the experiments, we use the initial version of MSR-VTT, referred to as MSR-VTT-10K, which contains 10K video clips from 20 categories. Each video clip is annotated with 20 sentences by 1,327 workers from Amazon Mechanical Turk. Therefore, the dataset results in a total of 200K clip-sentence pairs and 29,316 unique words.
We use the public splits for training and testing, \textit{i.e.}, 6,513 for training, 497 for validation, and 2,990 for testing.

\noindent \textbf{MSVD.} It contains 1,970 YouTube short video clips with each one depicting a single activity in 10 seconds to 25 seconds.
Each video clip has roughly 40 English descriptions. Similar to the prior work \cite{pan2016jointly,DBLP:conf/iccv/YaoTCBPLC15}, we take 1,200 video clips for training, 100 clips for validation and 670 clips for testing, respectively.

\noindent \textbf{ActivityNet.} It is a large-scale video benchmark dataset with the complex human activities for high-level video understanding tasks, including temporal action proposal, temporal action localization, and dense video captioning. Specifically, there are 10,024 videos for training, 4,926 for validation, and 5,044 for testing, respectively. Each video has multiple annotated events with starting and ending time-stamps as well as the corresponding captions.

\subsubsection{Implementation Details}

For the sentence preprocessing, we remove the punctuations in each sentence, split each sentence with blank space, and convert all words into lowercase.
The sentences longer than 30 are truncated, and the word embedding size for each word is set to 468. 

For the encoder, we feed all frames of each video clip into Inception-V4 \cite{DBLP:conf/aaai/SzegedyIVA17} which is pretrained on the ILSVRC-2012-CLS \cite{russakovsky2015imagenet} classification dataset for feature extraction after resizing them to the standard size of $299\times 299$, and extract the 1,536 dimensional semantic feature of each frame from the last pooling layer.
Inspired by \cite{DBLP:conf/iccv/YaoTCBPLC15}, we choose the equally-spaced 28 features from one video, and pad them with zero vectors if the number of features is less than 28. The input dimension of the decoder is 468, the same to that of the word embedding, while the hidden layer contains 512 units. For the reconstructor, the inputs are the hidden states of the decoder and thus have the dimension of 512. To ease the reconstruction loss computation, the dimension of the hidden layer is set to 1,536 same to that of the frame features produced by the encoder.

During the training, AdaDelta~\cite{zeiler2012adadelta} is employed for optimization with the cross-entropy loss while Adam~\cite{kingma2014adam} is applied with a learning rate 1e-5 under the RL. To help the models under the REINFORCE algorithm converge fast, we initialize them as the pre-trained models having the best CIDEr scores under cross entropy training. The training stops when the CIDEr value on the validation dataset stops increasing in the following 20 successive epochs. During the testing phase, beam search with size 5 is used for the final caption generation.

\subsection{Performance Comparisons}
\label{sec:study_ed}

\subsubsection{Performance on MSR-VTT}
\begin{table}[!t]
  \caption{Performance evaluation of different video captioning models on the testing set of the MSR-VTT dataset in terms of BLEU-4, METEOR, ROUGE-L, and CIDEr scores (\%). The encoder-decoder framework is equipped with different CNN structures such as AlexNet, GoogleNet, VGG19, and Inception-V4. Except Inception-V4, the metric values of the other published models are referred from the work in \cite{msr-vtt-large-video-description-dataset-bridging-video-language-supplementary-material}, and the symbol ``-'' indicates that such a metric is unreported.}
  \label{table1}
\scriptsize
  \centering
  \begin{tabular}{c|c|c|c|c}
  \hline
  Model                             & BLEU-4  &  METEOR  &  ROUGE-L  &  CIDEr \\ \hline \hline
  MP-LSTM (AlexNet)                 & 32.3    &  23.4    & -         &  -     \\
  MP-LSTM (GoogleNet)               & 34.6    &  24.6    &  -        &  -     \\
  MP-LSMT (VGG19)                   & 34.8    &  24.7    &  -        &  -     \\
  SA-LSTM (AlexNet)                 & 34.8    &  23.8    &  -        &  -     \\
  SA-LSTM (GoogleNet)               &  35.2   &  25.2    &  -        &  -     \\
  SA-LSTM (VGG19)                   & 35.6    &  25.4    &  -        &  -     \\  \hline
  SA-LSTM (Inception-V4)            & 36.3    &  25.5    &  58.3     &  39.9  \\  
  RecNet$_{global}$                 & 38.3    &  26.2    &  59.1     &  41.7  \\
  RecNet$_{local}$          & \textbf{39.1}    &  \textbf{26.6}    &  \textbf{59.3}     &  \textbf{42.7}  \\  \hline
  SA-LSTM (RL)               &  38.3   &   26.3   &   59.5    &  46.1  \\     
  RecNet(RL)$_{global}$      &   38.8  &   \textbf{27.5}   &   60.2   &  48.4  \\
  RecNet(RL)$_{local}$       &\textbf{39.2}   &\textbf{27.5}  &\textbf{60.3}  &\textbf{48.7}  \\  \hline
  \end{tabular}
\end{table}
In this section, we first test the impacts of different encoder-decoder architectures in video captioning, such as SA-LSTM and MP-LSTM. Both are popular encoder-decoder models and share similar LSTM structure, except that SA-LSTM introduced an attention mechanism to aggregate frame features, while MP-LSTM relies on the mean pooling. As shown in Table~\ref{table1}, with the same encoder VGG19, SA-LSTM yielded 35.6 and 25.4 on the BLEU-4 and METEOR metrics respectively, while MP-LSTM only produced 34.8 and 24.7, respectively. The same results can be obtained when using AlexNet and GoogleNet as the encoder. Hence, it is concluded that exploiting the temporal dynamics among frames with the attention mechanism performed better in sentence generation than mean pooling on the whole video.

\begin{figure} 
  \centering
  \vspace{7.5pt}
  \includegraphics[width=\hsize]{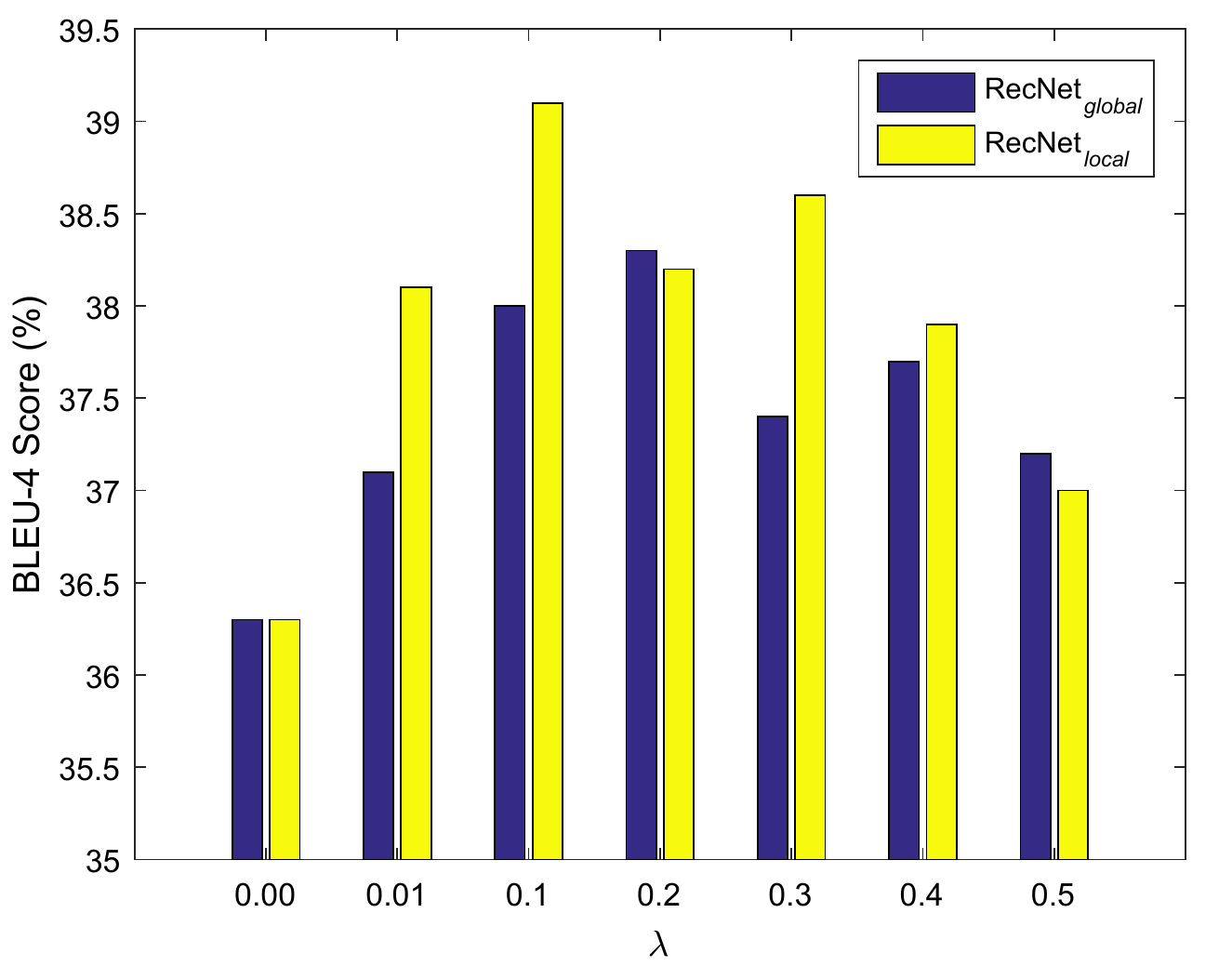} 
  \caption{Effects of the trade-off parameter $\lambda$ for RecNet$_{global}$ and RecNet$_{local}$ in terms of BLEU-4 metric on MSR-VTT. It is noted that $\lambda=0$ means that the reconstructor is off, and the RecNet turns to be a conventional encoder-decoder model.}
  \label{global_local_curve}
  \vspace{7.5pt}
\end{figure}

Besides, we also introduced Inception-V4 as an alternative CNN for feature extraction in the encoder. It is observed that with the same encoder-decoder structure SA-LSTM, Inception-V4 yielded the best captioning performance comparing to the other CNNs such as AlexNet, GoogleNet, and VGG19. This is probably because Inception-V4 is a deeper network and better at semantic feature extraction. Hence, SA-LSTM equipped with Inception-V4 is employed as the encoder-decoder model in the proposed RecNet.

By stacking the global or local reconstructor on the encoder-decoder model SA-LSTM, we can have the proposed encoder-decoder-reconstructor architecture: RecNet. Apparently, such a structure yielded significant gains to the captioning performance in all metrics. This proved that the backward flow information introduced by the proposed reconstructor could encourage the decoder to embed more semantic information and regularize the generated caption to be more consistent with the video contents. 
{
Actually, RecNet can be viewed as a dual learning model, where video captioning is the primal task and reconstruction behaves as its dual task. In the dual task, a reconstruction loss measuring the difference between the reconstructed and original visual features is employed additionally to train the primal task and optimize the parameters of the encoder and decoder. With such a reconstructor, the decoder is encouraged to embed more information from the input video sequence. Therefore, the relationship between the video sequence and caption can be enhanced. And the decoder can generate more semantically correlated sentences to describe the visual contents of the video sequences, yielding significant performance gains. More discussions and experimental results about the proposed reconstructor will be given in Section~\ref{sec:recon}.
}

Moreover, we compared RecNet with two models~\cite{jin2016describing,shetty2016frame}, which achieved great successes in 2016 MSR-VTT Challenge. As they utilized the multimodal features like aural and speech to boost captioning, which is beyond the focus of this work, we set the comparison on the C3D features only. Also, as they have reported results on the validation set of MSR-VTT, for a fair comparison we report out results on the same validation set. Our proposed RecNet$_{global}$ shows superiority over these two methods as in Table~\ref{table:c3d}. Incorporating the multimodal features into the proposed RecNet would be an interesting direction for us to pursue in the future.
\begin{table}
  \caption{Performance evaluation of different video captioning models with C3D features on the validation set of the MSR-VTT dataset in terms of BLEU-4, METEOR, ROUGE-L, and CIDEr scores (\%).}
  \label{table:c3d}
\scriptsize
  \centering
  \begin{tabular}{c|c|c|c|c}
  \hline
  Model                             & BLEU-4  &  METEOR  &  ROUGE-L  &  CIDEr \\ \hline \hline
  Qin \textit{et al.}~\cite{jin2016describing} &36.9& 27.3    & 58.4    & 41.9    \\
  Shetty \textit{et al.}~\cite{shetty2016frame} & 36.9    &  26.8    &  57.7    & 41.3  \\
  SA-LSTM                           & 37.9    &  26.9    &  59.5     &  41.5  \\  \hline
  RecNet$_{global}$                 & 38.5    &  27.5    &  59.7     &  42.7  \\ \hline
  \end{tabular}
\end{table}

Furthermore, we introduce the reinforcement learning into the SA-LSTM model and mix it with the proposed reconstructor. The results on MSR-VTT are shown in the bottom column of Table~\ref{table1}. It can be observed that REINFORCE algorithm makes a big contribution to the performance improvement of the baseline model, especially on the CIDEr value which is directly optimized by the model. Once again, we see the importance of backward flow mined by reconstructor: even if the network, such as the SA-LSTM(RL), has achieved great performance, collaborating with the backward information flow leads to a further improvement.

\subsubsection{Performance on MSVD}
Also, we tested the proposed RecNet on the MSVD dataset \cite{chen2011collecting}, and compared it to more benchmark encoder-decoder models, such as GRU-RCN\cite{ballas2015delving}, HRNE\cite{pan2016hierarchical}, h-RNN\cite{yu2016video}, LSTM-E\cite{pan2016jointly}, aLSTMs\cite{gao2017video}, and LSTM-LS\cite{liu2017video}. The quantitative results are given in Table~\ref{table2}. It is observed that the RecNet$_{local}$ and RecNet$_{global}$ with SA-LSTM performed the best and second best in all metrics, respectively. Besides, we introduced the reconstructor to S2VT\cite{DBLP:conf/iccv/VenugopalanRDMD15} to build another encoder-decoder-reconstructor model. The results show that both global and local reconstructors bring improvements to the original S2VT in all metrics, which again demonstrate the benefits of video captioning using bidirectional cue modeling.

\begin{table}[!h]
\caption{Performance evaluation of different video captioning models on the MSVD dataset
in terms of BLEU-4, METEOR, ROUGE-L, and CIDEr scores (\%). The symbol "-" indicates that such a metric is unreported.
}
  \label{table2}
  \scriptsize
  \centering
  \begin{tabular}{c|c|c|c|c}
  \hline
  Model                                              & BLEU-4  &  METEOR  &  ROUGE-L  &  CIDEr \\ \hline \hline
  MP-LSTM (AlexNet)\cite{venugopalan2015translating} & 33.3    &  29.1    & -         &  -     \\
  GRU-RCN\cite{ballas2015delving}					 & 43.3	   &  31.6    &  -		  &  68.0  \\
  HRNE\cite{pan2016hierarchical}					 & 43.8    &  33.1    &  -   	  &  -     \\
  LSTM-E\cite{pan2016jointly}                        & 45.3    &  31.0    &  -		  &  -     \\
  LSTM-LS (VGG19)\cite{liu2017video}                 & 46.5    &  31.2    &  -        &  -     \\
  h-RNN\cite{yu2016video}							 & 49.9    &  32.6    &  -        &  65.8  \\
  aLSTMs \cite{gao2017video}                         & 50.8    &  33.3    & -         &  74.8  \\  \hline
  S2VT (Inception-V4)                                & 39.6    &  31.2    &  67.5     &  66.7  \\
  SA-LSTM (Inception-V4)                             & 45.3    &  31.9    &  64.2     &  76.2  \\   \hline
  RecNet$_{global}$ (S2VT)                           & 42.9    &  32.3    &  68.5     &  69.3  \\
  RecNet$_{local}$  (S2VT)                           & 43.7    &  32.7    &  68.6     &  69.8  \\
  RecNet$_{global}$ (SA-LSTM)                        & 51.1    &  34.0    &  69.4     &  79.7  \\
  RecNet$_{local}$  (SA-LSTM)                 &\textbf{52.3} &\textbf{34.1}  &\textbf{69.8} &  \textbf{80.3}  \\   \hline
  \end{tabular}
\end{table}

One interesting point of the proposed reconstructor is about the capability of learning from limited amounts of data, which is brought by the dual learning mechanism. The benefit of RecNet on one limited training set can be demonstrated by comparing the results on MSVD and MSR-VTT. The gains on MSVD are more evident than those on MSR-VTT, while the size of MSVD is only one third of MSR-VTT. Taking the ResNet$_{local}$ as an example, we halve the size of MSVD to further demonstrate the proposed model on learning from limited data. As shown in Table~\ref{table:limited_data}, the performance on CIDEr declines when the data size is reduced. However, a larger improvement is achieved, with gain of 9.3\% on half of MSVD vs. 5.7\% on full MSVD.  

\begin{table}[!h]
  \caption{Performance on MSVD with reduced size in terms of CIDEr scores (\%). The RecNet$_{local}$ is token as the exemplary model.}
  \label{table:limited_data}
  \scriptsize
  \centering
  \begin{tabular}{c|c|c|c}
  \hline
  Data Size     & SA-LSTM (Inception-V4)  &  RecNet$_{local}$  &  Performance Gain   \\ \hline \hline
  50\%			& 72.2	& 78.9  & +9.3\%	\\
  100\%			& 76.2  & 80.3  & +5.4\%	\\ \hline
  \end{tabular}
\end{table}

\subsubsection{Performance on ActivityNet}
To further verify the effectiveness of the proposed RecNet, experiments on the ActivityNet dataset~\cite{caba2015activitynet} are conducted and shown in Table~\ref{table:activitynet_all}. 
We construct the video-sentence pairs by extracting the video segments indicated by the starting and ending timestamps as well as their associated sentences. As the ground-truth captions and timestamps of the test split are unavaliable, we validate our model on the validation split.

\begin{table}[!h]
  \caption{Performance evaluation of different video captioning models on the validation split of the ActivityNet dataset in terms of BLEU-4, METEOR, ROUGE-L, and CIDEr scores (\%). "RL" in brackets means that the model is trained by the self-critical algorithm.}
  \label{table:activitynet_all}
\scriptsize
  \centering
  \begin{tabular}{c|c|c|c|c} 
  \hline
  Model             & BLEU-4  &  METEOR  &  ROUGE-L  &  CIDEr \\ 
  \hline \hline
  SA-LSTM           &1.58  &8.69 & 18.24   &28.50   \\
  RecNet$_{global}$ &1.67  &9.23 &20.07  &34.78  \\
  RecNet$_{local}$  &1.71  &9.70 &20.50  &35.52  \\ 
  \hline
  SA-LSTM (RL)          &1.65  &9.37   &21.03    &33.57  \\     
  RecNet(RL)$_{global}$ &1.72  &10.43   &23.25  &37.96 \\
  RecNet(RL)$_{local}$  &1.74   &10.47  &23.49   &38.43  \\  
  \hline
  \end{tabular}
\end{table}

Similar to the results on MSR-VTT and MSVD, it can be observed that the proposed RecNet outperforms the base model SA-LSTM. The main reason can be attributed to that the backward flow is well captured by the proposed reconstructor. The same situation occurs when models are trained by the self-critical sequence training strategy. The methods realized in a traditional encoder-decoder architecture mainly focus on exploiting the forward flow (video to sentence), while neglect the backward flow (sentence to video). In contrast, the proposed RecNet, realized in a novel encoder-decoder-reconstructor architecture, leverages both the forward and backward flows. Therefore, the relationship between the video sequence and caption can be further enhanced, which can improve the video captioning performance.

\subsection{Study on the Trade-off Parameter $\lambda$}
In this section, we discuss the influence of the trade-off parameter $\lambda$ in Eq.~(\ref{eq_full_loss}). With different $\lambda$ values, the obtained BLEU-4 metric values are given in Fig.~\ref{global_local_curve}. First, it can be concluded again that adding the reconstruction loss ($\lambda>0$) did improve the performance of video captioning in terms of BLEU-4. Second, there is a trade-off between the forward likelihood loss and the backward reconstruction loss, as too large $\lambda$ may incur noticeable deterioration in the caption performance. Thus, $\lambda$ needs to be more carefully selected to balance the contributions of the encoder-decoder and the reconstructor. As shown in Fig.~\ref{global_local_curve}, we empirically set $\lambda$ to 0.2 and 0.1 for RecNet$_{global}$ and RecNet$_{local}$, respectively.

\subsection{Study on the Reconstructors}
\label{sec:recon}
A deeper study on the proposed reconstructor is discussed in this section.

\subsubsection{Performance Comparisons between Different Reconstructors} 
The quantitative results of RecNet$_{local}$ and RecNet$_{global}$ on MSR-VTT are given on the bottom two rows of Table~\ref{table1}. It can be observed that RecNet$_{local}$ performs slightly better than RecNet$_{global}$. The reason mainly lies in the temporal dynamic modeling. RecNet$_{global}$ employs mean pooling to reproduce the video representation and misses the local temporal dynamics, while the attention mechanism is included in RecNet$_{local}$ to exploit the local temporal dynamics for each frame reconstruction.

The performance gap on some metrics, such as METEOR and ROUGE-L, between RecNet$_{global}$ and RecNet$_{local}$ may be not significant. One possible reason is that the visual information of consecutive frames is very similar. As the video clips from the available datasets are short, the visual representations of frames are of small differences from each other. Therefore, the global and local structure information seems to be similar. Another possible reason is the complicated video-sentence relationship, which may lead to similar input information for RecNet$_{global}$ and RecNet$_{local}$.

\subsubsection{Effects of Reconstruction Order}
Another interesting point about the proposed reconstructor is that we do not need to constrain the reconstruction order. In fact, the reconstruction ordering is unnecessary when reproducing the global structure of the video. We may discard the idea in RecNet$_{global}$, as we finally put the mean pooling operation on the reconstructed feature sequence to acquire the video global structure (mean frame feature), which will disrupt the original feature order. For the RecNet$_{local}$, we need the correct reconstruction order and address it in an implicit manner, where the employed LSTM in RecNet reconstructs the video features frame by frame and then the local losses are pooled together as in Eq.~(\ref{eq_local_loss}). 

Besides, we have tested another method to constrain the correct reconstruction ordering for the RecNet$_{local}$. Specifically, we have tried replacing the attention mechanism in the reconstructor with the transposed attention matrix obtained in the decoder which has the ordering information about the input frame features. The results are given in Table~\ref{table:transposed}. It is observed that to a certain extent transposing the attention matrix for feature reconstruction does help boost the baseline encoder-encoder model. However, it is inferior to the proposed RecNet$_{local}$. The reason is that simply matrix transposing cannot effectively exploit the complicated relationship between the video features and the hidden states of the decoder.
\begin{table}
  \caption{Performance evaluation of different video captioning models on the testing set of the MSR-VTT dataset in terms of BLEU-4, METEOR, ROUGE-L, and CIDEr scores (\%).}
  \label{table:transposed}
\scriptsize
  \centering
  \begin{tabular}{c|c|c|c|c}
  \hline
  Model                             & BLEU-4  &  METEOR  &  ROUGE-L  &  CIDEr \\ \hline \hline
  SA-LSTM (Inception-V4)            & 36.3    &  25.5    &  58.3     &  39.9  \\
  RecNet$_{local}$                 & 39.1    &  26.6    &  59.3     &  42.7  \\  \hline
  RecNet$_{local}$(transpose)      & 36.7    &  25.7    &  57.9     &  40.4  \\ \hline
  \end{tabular}
\end{table}

\begin{table}[htbp]
    \caption{Performance evaluation of the combined architecture on MSR-VTT, MSVD, and ActivityNet in terms of BLEU-4, METEOR, and CIDEr scores (\%). }
    \label{table:global+local}
    \scriptsize
    \centering
    \begin{tabular}{c|c|c|c|c}
    \hline
    Dataset &Model &BLEU-4 &METEOR  &CIDEr \\
    \hline \hline
    \multirow{6}*{MSR-VTT}	&RecNet$_{global}$  & 38.3 &  26.2  &  41.7  \\
    ~       &RecNet$_{local}$ &39.1 &26.6  &42.3 \\ 
    ~		&RecNet$_{g+l}$ &38.7 &26.7 &43.1  \\
    ~       &RecNet(RL)$_{global}$      &   38.8  &   27.5   &  48.4  \\
    ~		&RecNet(RL)$_{local}$ &39.2 &27.5 &48.7  \\
    ~		&RecNet(RL)$_{g+l}$ & 39.3 &27.7 &49.5  \\
    \hline
    \multirow{6}*{MSVD}	&RecNet$_{global}$ & 51.1 & 34.0 &  79.7  \\
    ~       &RecNet$_{local}$ &52.3 &34.1 &80.3 \\
    ~		&RecNet$_{g+l}$ &51.5 &34.5 &81.8  \\
    ~       &RecNet(RL)$_{global}$ &52.3 &34.1 &82.9\\ 
    ~       &RecNet(RL)$_{local}$ &52.4 &34.3 &83.6  \\
    ~       &RecNet(RL)$_{g+l}$ &52.9 &34.8 &85.9  \\
    \hline
    \multirow{6}*{ActivityNet}	&RecNet$_{global}$ &1.67  &9.23 &34.78  \\
    ~       &RecNet$_{local}$ &1.71 &9.70 &35.52 \\
    ~		&RecNet$_{g+l}$ &1.72 &9.98 &35.84  \\
    ~       &RecNet(RL)$_{global}$ &1.72  &10.43  &37.96 \\
    ~       &RecNet(RL)$_{local}$ &1.74 &10.47 &38.43 \\
    ~		&RecNet(RL)$_{g+l}$ &1.75 &10.61 &38.88 \\
    \hline
    \end{tabular}
\end{table}
\subsubsection{Jointly Reconstructing the Global and Local Structures}

We also propose one new model named RecNet$_{g+l}$, which simultaneously considers both global and local structures to produce more reliable reconstruction of the video representation. The performances of RecNet$_{g+l}$ on MSR-VTT, MSVD, and ActivityNet are illustrated in Table~\ref{table:global+local}. Obviously, RecNet$_{g+l}$ yields consistent performance improvements over both RecNet$_{global}$ and RecNet$_{local}$, especially on the METEOR and CIDEr metrics.
The consistent performance improvement can also be observed when RecNets are trained by the self-critical strategy. For example, on the MSR-VTT dataset, RecNet(RL)$_{g+l}$ has a better CIDEr score (49.5) than RecNet(RL)$_{global}$ (48.4) and RecNet(RL)$_{local}$ (48.7), and the METEOR score of RecNet(RL)$_{g+l}$(27.7) also outperforms RecNet(RL)$_{global}$ (27.5) and RecNet(RL)$_{local}$ (27.5).
Such improvements on different benchmark datasets show that jointly learning the global and local reconstructors can help exploit the backward flow (sentence to video) comprehensively, thereby further boosting the performance of video captioning.

\subsubsection{Discussions on the Reconstructor}
\label{sec:curve_and_exposure}

\begin{figure} 
  \centering
  \includegraphics[width=\hsize]{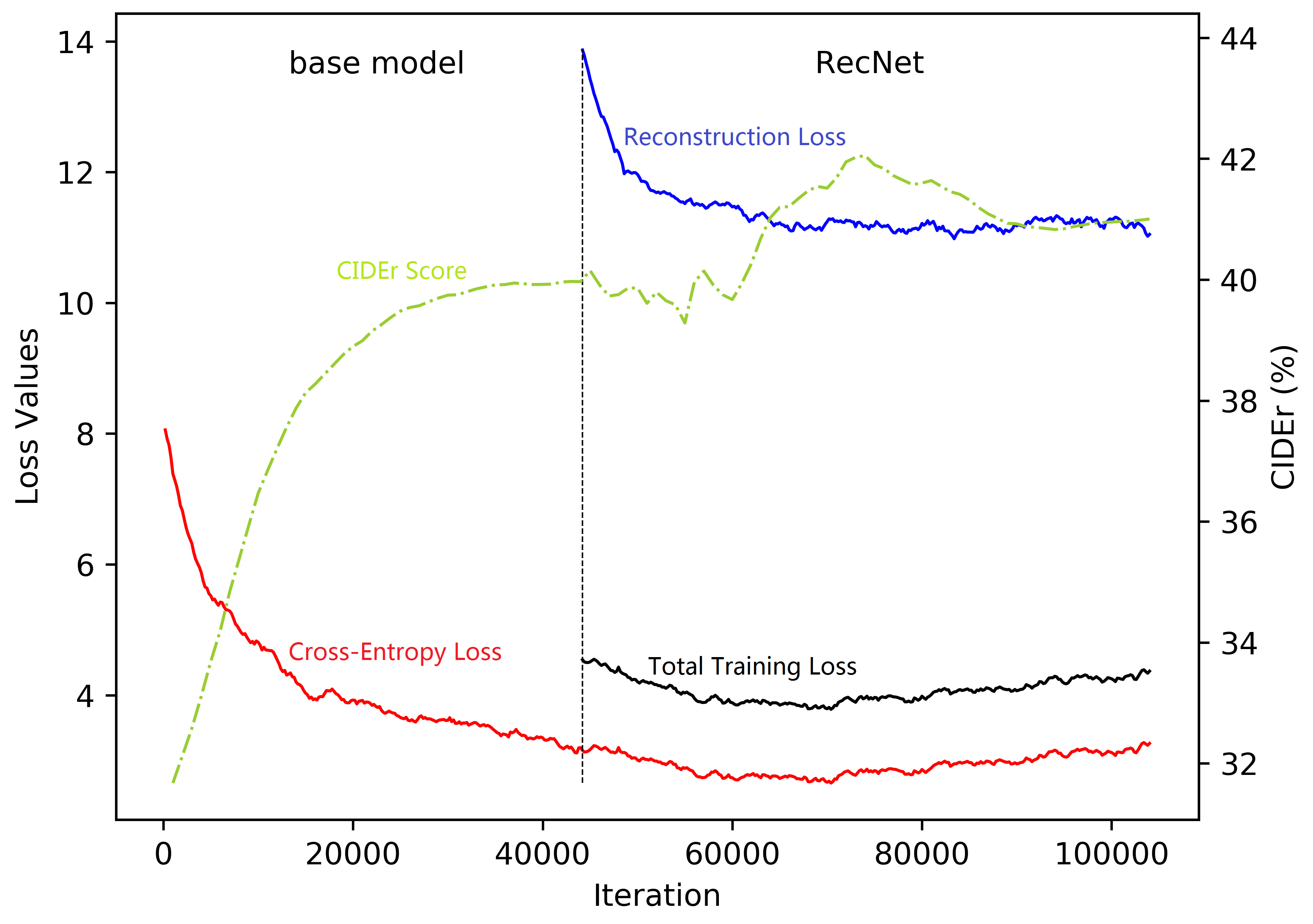} 
  \caption{
  The curves of training loss and CIDEr score during the training process (with and without the proposed reconstructor). The base model denotes the traditional encoder-decoder structure for video captioning, while the RecNet stacks the reconstructor on top of the encoder-decoder model. The solid lines in red, blue, and black indicate the cross-entropy loss, the reconstruction loss, and the total training loss, respectively. The dotted line in green indicates the CIDEr score.
  }
  \label{loss_curve}
\end{figure}

Moreover, we make more detailed studies on how well the reconstruction is performed.
First, we record the training losses as well as the CIDEr scores during the training process (with and without the proposed reconstructor), as illustrated in Fig.~\ref{loss_curve} to examine the effects of the reconstructor.

\begin{figure}
  \centering
  \includegraphics[width=\hsize]{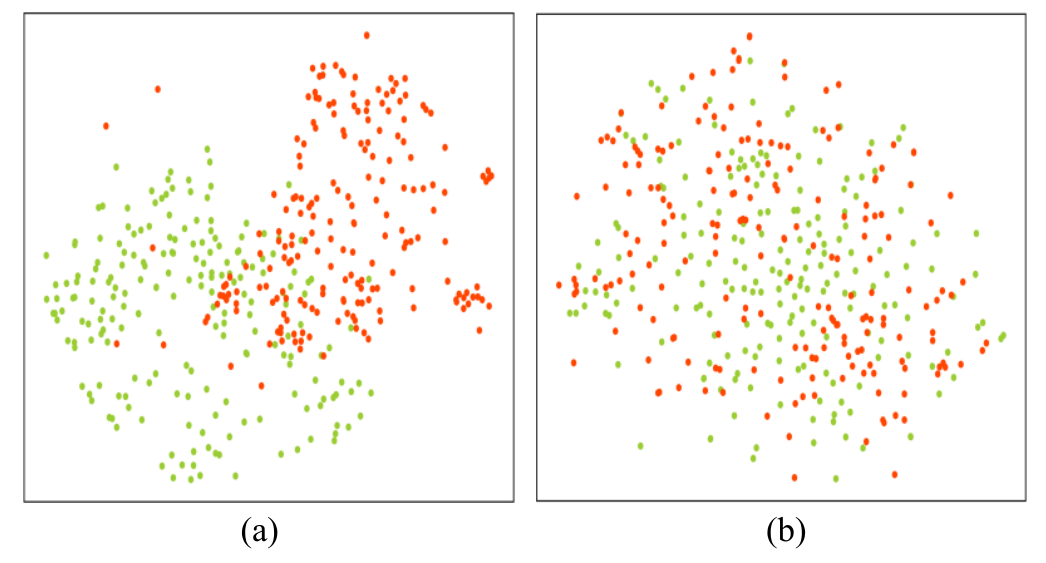} 
  \caption{Visualization of the distribution of the decoder hidden states in base model (a) and RecNet$_{local}$ (b). The dots in red represent the hidden states generated in training process, and dots in green represent the hidden states generated during the inference process.}
  \label{hidden_distribution}
\end{figure}

\begin{figure*}[htbp!]
  \centering
  \vspace{7.5pt}
  \includegraphics[width=0.97\hsize]{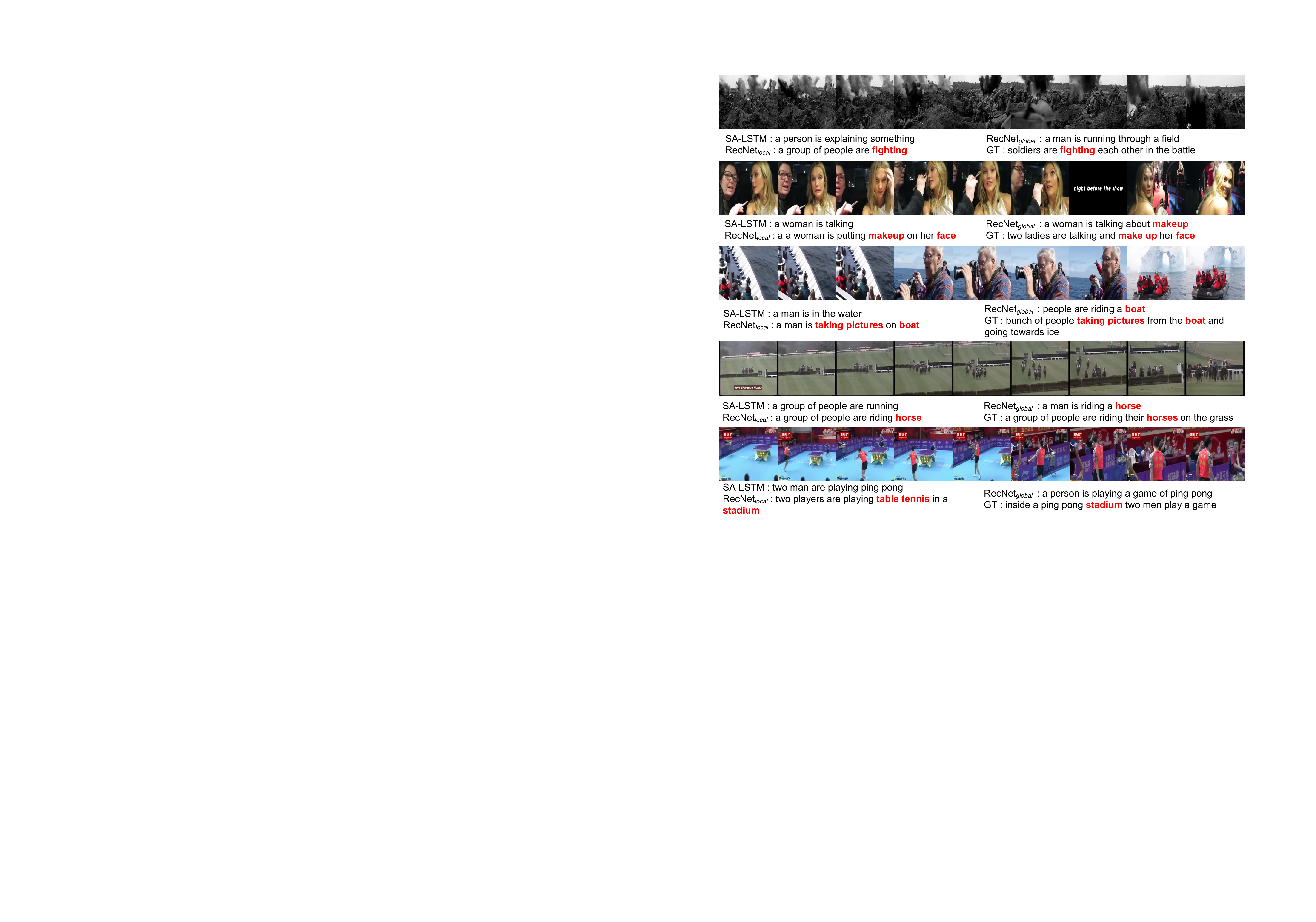} 
  \caption{Visualization of some video captioning examples on the MSR-VTT dataset with different models. Due to the page limit, only one ground-truth sentence is given as the reference. Compared to SA-LSTM, the proposed RecNet is able to yield more vivid and descriptive words highlighted in red boldface, such as ``\texttt{fighting}'', ``\texttt{makeup}'', ``\texttt{face}'', and ``\texttt{horse}''.}
  \label{samples}
  \vspace{7.5pt}
\end{figure*}

It can be observed that the cross-entropy loss defined by Eq.~(\ref{eq:XE_encoder-decoder}) and the CIDEr score consistently converge during the training process of the baseline model. Relying on the early stopping strategy, the parameters of the baseline model are determined. Afterward, the reconstructor, specifically the RecNet$_{local}$, is employed to reconstruct the local structure of the original video from the hidden state sequence generated by the decoder. During such a process, the reconstruction loss defined by Eq.~(\ref{eq_local_loss}) is used to train the reconstructor and the total training loss defined by Eq.~(\ref{eq_full_loss}) is used to train the encoder and decoder. It can be observed that the reconstruction loss decreases as the training proceeds, while the cross-entropy loss increases slightly in the beginning and then consistently decreases. It is reasonable as the untrained reconstructor can hardly provide valid information to train the encoder and decoder in the first few iterations. The same observations can be obtained for the CIDEr score. Hence, it can be concluded that the proposed reconstructor can help train the encoder and decoder and thereby improves the video captioning performance. Also, the decrease of the reconstruction loss clearly demonstrates that such reconstruction can reproduce meaningful visual information from the hidden state sequence generated by the decoder.

Second, we examine how the hidden state in the decoder changes after the reconstructor is employed. 
Generally, in the video captioning models trained with the cross-entropy loss, the decoder is encouraged to maximize the likelihood of each word given the previous hidden state and the ground-truth word at the previous step. Whereas for inference, the errors are accumulated along the generated sentence, as no ground-truth word is observed by the decoder before predicting a word. As such, the discrepancy between training and inference is inevitably introduced, which is also referred to as the exposure bias~\cite{ranzato2015sequence}.

To examine how the hidden state in the decoder changes after the reconstruction is introduced. We consider the distributions of the last hidden states of the sequence in the decoder as in~\cite{chen2018regularizing}, since they encode the necessary information about the whole sequential input. Specifically, we visualize the distributions of the last hidden states when the LSTM in the decoder reaches the maximum step or `$<\text{EOS}>$'~(the end signal of the sentence) is predicted, in the training and inference stages, respectively. The hidden state visualizations are illustrated in Fig.~\ref{hidden_distribution}. The hidden states are from the same batch with size 200. We reduce the dimension of the hidden states to 2 with T-SNEs~\cite{maaten2008visualizing}. It is obvious that the baseline model (Fig.~\ref{hidden_distribution}~(a)) in training and inference stages presents a large discrepancy, while the proposed RecNet (Fig.~\ref{hidden_distribution}~(b)) significantly reduces such a discrepancy. This is also one of the reasons that RecNet performs better than the competitor models.

\subsection{Qualitative Analysis}
Besides, some qualitative examples are shown in Fig.~\ref{samples}. Still, it can be observed that the proposed RecNets with local and global reconstructors generally produced more accurate captions than the typical encoder-decoder model SA-LSTM. For example, in the second example, SA-LSTM generated ``\texttt{a woman is talking}'', which missed the core subject of the video, \textit{i.e.}, ``\texttt{makeup}''. By contrast, the captions produced by RecNet$_{global}$ and RecNet$_{local}$ are ``\texttt{a woman is talking about makeup}'' and  ``\texttt{a women is putting makeup on her face}'', which apparently are more accurate. RecNet$_{local}$ even generates the word of ``\texttt{face}'', which results in a more descriptive caption. More results can be found in the supplementary file.

\section{Conclusions}
In this paper, we proposed a novel RecNet in the encoder-decoder-reconstructor architecture for video captioning, which exploits the bidirectional cues between natural language description and video content. Specifically, to address the backward information from description to video, two types of reconstructors were devised to reproduce the global and local structures of the input video, respectively. An additional architecture by fusing the two types of reconstructors was also presented and compared with the models that can reproduce either the global or the local structure separately. The forward likelihood and backward reconstruction losses were jointly modeled to train the proposed network. Besides, we employed the REINFORCE algorithm to directly optimize the CIDEr score and fused the reward-based loss with the traditional loss from the reconstructor for further improving the captioning performance. The extensive experimental results on the benchmark datasets demonstrate the superiority of the proposed RecNet over the existing encoder-decoder models in terms of the typical metrics for video captioning.

\section*{Acknowledgments}
This work was supported by the National Key Research and Development Plan of China under Grant 2017YFB1300205, NSFC Grant no. 61573222, and Major Research Program of Shandong Province 2018CXGC1503.


\ifCLASSOPTIONcaptionsoff
  \newpage
\fi

\bibliographystyle{IEEEtran}
\bibliography{IEEEabrv,mytipbib}

\end{document}